\documentclass[twoside]{article}
\usepackage{color}
\usepackage{amsmath,amsthm, amssymb,ascmac}
\usepackage{bm}
\usepackage{rotating}
\usepackage[numbers]{natbib}
\usepackage{algorithmic,algorithm}
\usepackage{url}

\newtheorem{thm}{Theorem}

\newtheorem{proposition}[thm]{Proposition}

\newtheorem{definition}[thm]{Definition}
\newtheorem{remark}[thm]{Remark}
\newtheorem{example}[thm]{Example}

\newtheorem{exercise}[thm]{Exercise}
\newtheorem{report}[thm]{Report}

\newcommand{\RR}{\mathbb{R}}
\newcommand{\ZZ}{\mathbb{Z}}

\newcommand{\cF}{{\mathcal F}}

\newcommand{\Ba}{\bm{a}}
\newcommand{\tBa}{\tilde{\bm{a}}}

\newcommand{\Be}{\bm{e}}
\newcommand{\Bf}{\bm{f}}

\newcommand{\BR}{\bm{R}}
\newcommand{\Bs}{\bm{s}}

\newcommand{\By}{\bm{y}}
\newcommand{\Bbeta}{\bm{\beta}}
\newcommand{\Bepsilon}{\bm{\epsilon}}
\newcommand{\Bgamma}{\bm{\gamma}}

\newcommand{\abs}[1]{\vert{#1}\vert}

\newcommand{\norm}[1]{\vert\vert #1 \vert\vert}

\newcommand{\argmin}{\mathop{\mathrm{argmin}}}

\newcommand{\Var}{\mathrm{Var}}
\newcommand{\st}{\mathrm{s.t.}}
\newcommand{\dist}{\mathop{\mathrm{dist}}}

\title{\sc The Group Lasso for Design of Experiments}
\author{
Kentaro TANAKA \\
{\small Faculty of Economics, } \\
{\small Seikei University.} \\
{\small 3-3-1, Kichijoji-Kitamachi, Musashino-shi, Tokyo, 180-8633, Japan} \\
{\small e-mail: tanaken@econ.seikei.ac.jp}
\and
Masami MIYAKAWA \\
{\small Department of Industrial Engineering and Management, } \\
{\small Graduate School of Decision Science and Technology, } \\
{\small Tokyo Institute of Technology.} \\
{\small 2-12-1, O-okayama, Meguro-ku, Tokyo 152-8552, Japan.} \\
{\small e-mail: miyakawa.m.aa@m.titech.ac.jp}
}
\newfont{\m}{cmr8}
\newfont{\ms}{cmsl8}
\begin{document}
\maketitle
\begin{abstract}
We introduce an application of the group lasso to design of experiments.
Note that we are NOT trying to explain experimental design for the group lasso.
Conversely, we explain how we can use the idea of the group lasso in experimental design,
showing that the problem of constructing an optimal design matrix
can be transformed into a problem of the group lasso.
In some numerical examples, we show
that we can obtain the orthogonal arrays as the solutions of the group lasso problems.
\end{abstract}

\begin{flushleft}
\end{flushleft}

\section{Introduction}
\label{sec:introduction}

Design of experiments is widely used in a variety of fields
such as agriculture, quality control, and simulation.
One of the purpose of design of experiments is to construct the optimal experimental design
with respect to a criterion under some constraints reflecting real problem.
However, it is sometimes hard to obtain the optimal designs theoretically.
Recently, several computer-based approaches have been developed for this problem.
Especially,
mixed integer programming is used to
construct balanced incomplete block designs (\citet{YokoyaYamada2011}),
orthogonal designs (\citet{VieiraSanchezKienitzBelderrain2011a})
and
nearly orthogonal nearly balanced mixed designs (\citet{VieiraSanchezKienitzBelderrain2011b}).
Additionally, \citet{JonesNachtsheim2011} gives randomized algorithm to obtain preferable designs
for screening with factors having three levels.
Interestingly, in \citet{XiaoLinBai2012}, they theoretically give the optimal designs for the same problem of \citet{JonesNachtsheim2011},
using conference matrices (\citet{Belevitch1950}).

In this paper, we propose a new machine learning approach to obtain
an optimal experimental design.
In our approach, we consider that a design which has fewer design points and smaller variances for what we want to estimate is preferable.
In general, the fewer the design points, the larger the variances of estimates.
Therefore, there exists a trade-off between the number of design points and the variances of estimators.

First, to construct an optimal design, we enumerate the candidate design points.
In many cases, it is difficult to conduct the experiments at every candidate design points from the aspect of the cost.
Therefore, it is necessary to choose the subset of design points among them with minimum loss of information.
In Section~\ref{subsec:DOE}, we explain that this procedure corresponds to the method of variable selection for regression analysis
and can be formulated as the problem of the group lasso.

The organization of this paper is as follows.
In Section~\ref{sec:application-of-group-lasso-to-design-of-experiments},
we review the formulation of the group lasso (Section~\ref{subsec:group-lasso})
and apply it to the problem of constructing an optimal design (Section~\ref{subsec:DOE}).
Section~\ref{sec:numerical-examples} is devoted to the numerical examples of our approach.
In Section~\ref{sec:concluding-remarks}, we summarize the features of our approach.

\section{An application of the group lasso to design of experiments}
\label{sec:application-of-group-lasso-to-design-of-experiments}
\subsection{The group lasso}
\label{subsec:group-lasso}

The method of the group lasso (\citet{YuanLin2007}) which is a kind of generalization of the lasso (\citet{Tibshirani1996})
has become a popular method of variable selection for linear regression.
Let us consider the usual linear regression:
we have continuous outputs $\By \in \RR^{N}$ and a $N \times D$ design matrix $X$,
where $N$ is the sample size and $D$ is the number of input variables.
The estimator of the group lasso $\hat{\Bbeta} \in \RR^{D}$ is defined as
\begin{eqnarray}
 \hat{\Bbeta} = \argmin_{\Bbeta \in \RR^{D}}
  \norm{\By - X \Bbeta}_{2}^{2} + \sum_{g=1}^{G} \lambda_{g} \norm{\Bbeta_{I_{g}}}_{2},
  \label{eq:group-lasso}
\end{eqnarray}
where
$\lambda_{1}, \dots, \lambda_{G}$ are tuning parameters,
$\norm{\cdot}_{2}$ stands for the Euclidean norm (not squared),
$I_{1}, \dots, I_{G}$ are disjoint subsets of $\{1,\dots,D\}$
and
$\Bbeta_{I_{g}}=(\beta_{i_{1}}$, $\dots$, $\beta_{i_{g}})$
for $I_{g}=\{i_{1},\dots,i_{g}\}$.
If $I_{1}, \dots, I_{G}$ are all singletons, then the group lasso of \eqref{eq:group-lasso} coincides with the lasso.
The group lasso has the property that it does variable selection at the group level, i.e.,
an entire group of input variables may drop out of the model.
The group lasso problem of \eqref{eq:group-lasso} can be formulated as a second order cone programming and solved by the interior point methods.
Furthermore, there are some specialized algorithms which solve the group lasso problem faster than the interior point methods.

\subsection{Design of experiments}
\label{subsec:DOE}

Assume that there are $F$ factors $a_{1},\dots,a_{F}$ and the relation between the response variable $R$ and the factors is formulated as
\begin{eqnarray}
 R = \sum_{\Bf \in \cF} \gamma_{\Bf}\Ba^{\Bf} + \epsilon,
  \label{eq:model}
\end{eqnarray}
where $\cF$ is a finite subset of $\ZZ_{\geq 0}^{F}$,
$\Ba^{\Bf}=\prod_{i=1}^{F}a_{i}^{f_{i}}$ for $\Bf=(f_{1},\dots,f_{F}) \in \cF$
and
$\epsilon$ is the error with mean zero and variance $\sigma^{2}$.
Here the set of the parameters which we want to estimate is a subset of $\{\gamma_{\Bf} \in \RR \mid \Bf \in \cF\}$.
For simplicity, in this paper, we only consider the case where each factor has finite fixed levels
and assume that the $f$-th factor has $l_{f}$ levels $a_{f1},\dots,a_{fl_{f}}$.
Let $A$ be the set of all candidate design points.
If we consider all combinations of the levels of the factors without repetition as the candidate design points,
then $A=\{(a_{1i_{1}},\dots,a_{Fi_{F}})^{T} \mid i_{f} \in \{1,\dots,l_{f}\}\}$
where $T$ stands for transpose and $A$ has $\prod_{f=1}^{F}l_{f}$ elements.
Let $G$ be the number of elements of $A$ and $A=\{\Ba_{1},\dots,\Ba_{G}\}$ sorting the elements of $A$, i.e. the candidate design points, in an arbitrary order.
The design matrix $C$ of the candidate points is defined such that
the $g$-th column is $\Ba_{g}$, i.e., $C = [\Ba_{1}\,\dots\,\Ba_{G}]$.
\begin{example}
\label{example:F3Level2}
 Let us consider the case where $F=3$ and each factor has two levels $1$ or $-1$.
If we consider all combinations of the levels of the factors without repetition as the candidate design points,
then there are $G=2^{3}=8$ candidate design points and a design matrix of them is defined as follows:
\begin{eqnarray}
C =
 \begin{bmatrix}
 1 &  1 &  1 &  1 & -1 & -1 & -1 & -1 \\
 1 &  1 & -1 & -1 &  1 &  1 & -1 & -1 \\
 1 & -1 &  1 & -1 &  1 & -1 &  1 & -1
 \end{bmatrix}
 .
 \label{eq:F3Level2:main}
\end{eqnarray}
\end{example}

Let $\abs{\cF}$ be the number of elements in $\cF$
and $\cF=\{\Bf_{1},\dots,\Bf_{\abs{\cF}}\}$ sorting the elements of $\cF$ in an arbitrary order.
For each $\Ba_{g}$, let $\tBa_{g}=(\Ba_{g}^{\Bf_{1}},\dots,\Ba_{g}^{\Bf_{\abs{\cF}}})^{T}$.
Then a model matrix $M$ of the candidate points is defined such that
the $g$-th column is $\tBa_{g}$, i.e., $M = [\tBa_{1}\,\dots\,\tBa_{G}]$.
\begin{example}
 \label{example:F3Level2:2}
 As in Example~\ref{example:F3Level2},
we consider the case where $F=3$ and each factor has two levels $1$ or $-1$.
Furthermore, we assume that the relation between the response variable and the factors is formulated as
\begin{eqnarray}
 R = \gamma_{000} + \gamma_{100}a_{1} + \gamma_{010}a_{2} + \gamma_{001}a_{3} + \epsilon.
  \label{eq:F3Level2:model-eq}
\end{eqnarray}
This is called the main effect model.
Then
$\cF=\{(0,0,0), (1,0,0), (0,1,0), (0,0,1)\}$ and
a model matrix of candidate design points is defined as follows:
\begin{eqnarray}
M =
 \begin{bmatrix}
  1 &  1 &  1 &  1 &  1 &  1 &  1 &  1 \\
  1 &  1 &  1 &  1 & -1 & -1 & -1 & -1 \\
  1 &  1 & -1 & -1 &  1 &  1 & -1 & -1 \\
  1 & -1 &  1 & -1 &  1 & -1 &  1 & -1
 \end{bmatrix}
 .
 \label{eq:F3Level2:model-mat}
\end{eqnarray}
Note that, for simplicity, we may also write $\gamma_{0}, \gamma_{1}, \gamma_{2}$ and $\gamma_{3}$
instead of $\gamma_{000}$, $\gamma_{100}$, $\gamma_{010}$ and $\gamma_{001}$.
\end{example}

Let $\Be_{j}=\Be_{\Bf_{j}}$ be the $\abs{\cF}$-dimensional column vector such that
$\Be_{j}$ has an entry $1$ in the row corresponding to $j$ and $0$ otherwise.
Furthermore,
for $j \in \{1,\dots,\abs{\cF}\}$,
let $\gamma_{j}=\gamma_{\Bf_{j}}$ as in Example~\ref{example:F3Level2:2},
and $\Bgamma=(\gamma_{1},\dots,\gamma_{\abs{\cF}})^{T}$.
Let $R_{g}$ and $\epsilon_{g}$ be the response variable and the error respectively
when the factors $\Ba_{g}^{\Bf_{1}},\dots,\Ba_{g}^{\Bf_{\abs{\cF}}}$ are given.
Let $\BR=(R_{1},\dots,R_{G})^{T}$ and $\Bepsilon=(\epsilon_{1},\dots,\epsilon_{G})^{T}$.
We use the following proposition
to formulate the problem of constructing an optimal design matrix
as a problem of mathematical programming.
\begin{proposition}
 \label{prop:unbiasedness-and-variances}
 If there exists $\Bbeta_{j}=(\beta_{j1},\dots,\beta_{jG}$) such that
 $\beta_{j1}\tBa_{1}+\dots+\beta_{jG}\tBa_{G}=\Be_{j}$,
 then $\hat{\gamma_{j}}=\beta_{j1}R_{1}+\dots+\beta_{jG}R_{G}$ is an unbiased estimator of $\gamma_{j}$.
 The variance of $\hat{\gamma_{j}}$ is $\Var[\hat{\gamma_{j}}] = \sigma^{2} \norm{\Bbeta_{j}}_{2}^{2}$.
\end{proposition}
\begin{proof} 
Note that $\BR=M^{T}\Bgamma+\Bepsilon$ and $M\Bbeta_{j}=\Be_{j}$ from the assumption.
Then we obtain
\begin{eqnarray}
 E[\hat{\gamma_{j}}]
  &=& E[\beta_{j1}R_{1}+\dots+\beta_{jL}R_{G}]
  \; =\; E[\BR^{T}\Bbeta_{j}]
  \; =\; E[(M^{T}\Bgamma+\Bepsilon)^{T}\Bbeta_{j}]
  \nonumber \\
  &=&E[\Bgamma^{T} M\Bbeta] + E[\Bepsilon^{T}\Bbeta_{j}]
  \; =\; E[\Bgamma^{T} \Be_{j}]
  \; =\; \gamma_{j}.
  \nonumber
\end{eqnarray}
The variance is
\begin{eqnarray}
 \Var[\hat{\gamma_{j}}]
  \; = \;
  \beta_{j1}^{2} \Var[R_{1}]+\dots+\beta_{jL}^{2}\Var[R_{G}]
  \; = \; \sigma^{2}\norm{\Bbeta_{j}}_{2}^{2}.
  \nonumber
\end{eqnarray}
\end{proof}

As our design criterion, we use the A-optimality in which
the sum (or the ``A''verage) of the variances of the estimators is minimized.
At the same time, we consider the problem to choose the bare minimum of design points to save the cost.
Assume that the parameters in the set $\Bgamma_{J}=\{\gamma_{j}\mid j \in J\}$ are what we want to estimate where $J \subseteq \{1,\dots,\abs{\cF}\}$.
For each $\gamma_{j} \in \Bgamma_{J}$, we consider a linear estimator $\hat{\gamma_{j}}=\beta_{j1}R_{1}+\dots+\beta_{jG}R_{G}=\BR^{T}\Bbeta_{j}$.
From Proposition~\ref{prop:unbiasedness-and-variances},
the sum of the variances of the estimators for $\{\gamma_{j}\mid j \in J\}$ is $\sum_{j \in J} \norm{\Bbeta_{j}}_{2}^{2}$.
Then based on the A-optimality criterion, we need to minimize the sum $\sum_{j \in J} \norm{\Bbeta_{j}}_{2}^{2}$
under the condition of the unbiasedness: $M\Bbeta_{j}=\Be_{j}$ for $j \in J$.
Next, we consider the sparseness of the estimator.
Let
$\Bbeta_{I_{g}}$ be a column vector whose entries are $\{\beta_{jg} \mid j \in J\}$.
If $\Bbeta_{I_{g}}=\bm{0}$, then the response $R_{g}$ at the $g$-th design point is not used for the estimators $\{\hat{\gamma_{j}} \mid j \in J\}$.
Therefore, using the spirit of the group lasso, the following penalized least square gives the solution
which minimizes the sum of the variances of the estimators under the condition of the unbiasedness
with consideration of the sparseness.
\begin{eqnarray}
 \begin{array}{ccc}
 \displaystyle{\min_{\{\Bbeta_{j}\mid j \in J\}}} & & \displaystyle{\sum_{j \in J} \norm{\Bbeta_{j}}_{2}^{2} + \sum_{g=1}^{G} \lambda_{g}\norm{\Bbeta_{I_{g}}}}
  \\ & & \\
 \st & & M\Bbeta_{j}=\Be_{j}, \quad (j \in J).
 \end{array}
 \label{eq:GroupLasso-SOCP-DOE}
\end{eqnarray}
Here, $\lambda_{1}, \dots, \lambda_{G}$ are tuning parameters.
The problem of \eqref{eq:GroupLasso-SOCP-DOE} is a second order cone programming and can be solved by the interior point methods.
Note that the formulation of \eqref{eq:GroupLasso-SOCP-DOE} contains linear constraints and
some specialized algorithms which solves the group lasso problem may not applicable.
We can also consider the Lagrangian relaxation problem of \eqref{eq:GroupLasso-SOCP-DOE} as follows:
\begin{eqnarray}
 \min_{\{\Bbeta_{j}\mid j \in J\}} & & \sum_{j \in J}
  (\norm{\Bbeta_{j}}_{2}^{2} + \kappa_{j}\norm{M\Bbeta_{j}-\Be_{j}}_{2}^{2})
  +
  \sum_{g=1}^{G} \lambda_{g}\norm{\Bbeta_{I_{g}}}
 \label{eq:GroupLasso-Relaxation-DOE}
\end{eqnarray}
where, $\kappa_{j},\, (j \in J)$ is a tuning parameter.
The formulation of \eqref{eq:GroupLasso-Relaxation-DOE} is the same
as the group lasso and thus the specialized algorithms for the group lasso is applicable.
In particular, the solution of the problem of \eqref{eq:GroupLasso-Relaxation-DOE} does not accurately satisfy
the constraints of the unbiasedness when $\kappa_{j}$'s are not so large.
This means that, from the problem of \eqref{eq:GroupLasso-Relaxation-DOE},
we can obtain the solution when we allow confounding among the factors.
Therefore, the formulation of \eqref{eq:GroupLasso-Relaxation-DOE} works
even when the number of the elements of $J$ is larger than $G$ or the number of non-zero $\Bbeta_{I_{g}}$'s.

As the following example shows, in the formulation of \eqref{eq:GroupLasso-SOCP-DOE} or \eqref{eq:GroupLasso-Relaxation-DOE},
we have to determine the values of
$\lambda_{g}$'s and $\kappa_{j}$'s
carefully to obtain the sparse solution.
\begin{example}
 \label{example:symmetry}
 We assume that
 there are $F=3$ factors and
 each factor has two levels $1$ or $-1$.
 Furthermore, we assume that the candidate design points consist of all combinations of the levels of the factors as in \eqref{eq:F3Level2:main}.
 We consider the main effect model: $R=\gamma_{0}+\sum_{j=1}^{3}\gamma_{j}a_{j}+\epsilon$.
 Then the model matrix is given by \eqref{eq:F3Level2:model-mat}.
 Suppose that we want to estimate $\gamma_{1}$, $\gamma_{2}$ and $\gamma_{3}$.
 \begin{table}[htpb]
  \begin{minipage}{0.5\hsize}
   \begin{center}
    \caption{$L_{4}$ orthogonal array}
    \label{table:L4}
    \begin{tabular}{c|cccc}
     Run & $1$ & $2$ & $3$ & $4$ \\
     \hline
     $a_{1}$ & 1 & 1 & -1 & -1 \\
     $a_{2}$ & 1 & -1 & 1 & -1 \\
     $a_{3}$ & 1 & -1 & -1 & 1
    \end{tabular}
   \end{center}
  \end{minipage}
  \begin{minipage}{0.5\hsize}
   \begin{center}
    \caption{The orthogonal array $(-1) \times L_{4}$}
     \label{table:L4inverted}
    \begin{tabular}{c|cccc}
     Run & $1$ & $2$ & $3$ & $4$ \\
     \hline
     $a_{1}$ & -1 & -1 & 1 & 1 \\
     $a_{2}$ & -1 & 1 & -1 & 1 \\
     $a_{3}$ & -1 & 1 & 1 & -1
    \end{tabular}
   \end{center}
 \end{minipage}
 \end{table}
 To this model, traditional design of experiments shows that $L_{4}$ orthogonal array in Table~\ref{table:L4} is optimal.
 Note that Table~\ref{table:L4} is transposed against the traditional notation of $L_{4}$ orthogonal array,
 i.e., each column corresponds to a run of the experiment and the rows indicate the levels of $a_{1}$, $a_{2}$ and $a_{3}$.
 The feasible solution of \eqref{eq:GroupLasso-SOCP-DOE} which corresponds to $L_{4}$ orthogonal array is
 \begin{eqnarray}
  \Bbeta_{1} &=& \frac{1}{4}(0,1,1,0,-1,0,0,-1)^{T}, \nonumber \\
  \Bbeta_{2} &=& \frac{1}{4}(0,1,-1,0,1,0,0,-1)^{T},
  \label{eq:L4beta}  \\
  \Bbeta_{3} &=& \frac{1}{4}(0,1,-1,0,-1,0,0,1)^{T}. \nonumber
 \end{eqnarray}
 From Gauss-Markov theorem, the feasible solution \eqref{eq:L4beta} can be obtained by least squares of $\gamma_{1}$, $\gamma_{2}$ and $\gamma_{3}$
 for the experiment which has exactly four design points:
 $2$nd, $3$rd, $5$th and $8$th columns of \eqref{eq:F3Level2:main}.
 However, as shown below, if we set $\lambda_{1}=\dots=\lambda_{8}=\lambda$ for any non-negative real number $\lambda$,
 then we can not obtain $L_{4}$ orthogonal array as the optimal solution of \eqref{eq:GroupLasso-SOCP-DOE}.
 Let us consider the following feasible solution:
 \begin{eqnarray}
  \Bbeta_{1} &=& \frac{1}{4}(-1,0,0,-1,0,1,1,0)^{T}, \nonumber \\
  \Bbeta_{2} &=& \frac{1}{4}(-1,0,0,1,0,-1,1,0)^{T},
  \label{eq:L4beta2} \\
  \Bbeta_{3} &=& \frac{1}{4}(-1,0,0,1,0,1,-1,0)^{T}. \nonumber
 \end{eqnarray}
 This solution corresponds to the orthogonal array in Table~\ref{table:L4inverted}.
 It can be easily seen that the value of the objective function of \eqref{eq:GroupLasso-SOCP-DOE} for \eqref{eq:L4beta}
 and that for \eqref{eq:L4beta2} are the same.
 Furthermore, note that the problem of \eqref{eq:GroupLasso-SOCP-DOE} is strictly convex and thus the minimum is attained at a unique point.
 Therefore, the feasible solution \eqref{eq:L4beta} is not the optimal solution of \eqref{eq:GroupLasso-SOCP-DOE}.
\end{example}
The above example shows that the presence of the symmetries in the formulation of \eqref{eq:GroupLasso-SOCP-DOE} or \eqref{eq:GroupLasso-Relaxation-DOE}
prevents the sparseness of the solution.
In Section \ref{sec:numerical-examples}, we give a method to construct $\lambda_{1},\dots,\lambda_{G}$
which produces asymmetries in \eqref{eq:GroupLasso-SOCP-DOE}
and enables us to obtain the orthogonal arrays in several situations.

\section{Numerical examples}
\label{sec:numerical-examples}

In this section, we mainly consider the case where each factor has two levels $\{+1, -1\}$
and the candidate design points consist of all combinations of the levels of the factors.
We denote the set of candidate design points by $C=[\Ba_{1} \dots \Ba_{G}]$ as in Section \ref{subsec:DOE},
and assume $\Ba_{1}=(-1,\dots,-1)^{T}$.
For our experiment, a laptop with an 1.20 GHz CPU and 8GB RAM is used.
You can find the file of the program written in Python used in the following examples at \url{https://github.com/tanaken-basis/explasso}.

As Example~\ref{example:symmetry} shows, we need to carefully determine the values of $\lambda_{1},\dots,\lambda_{G}$
to obtain the sparse solution.
In the following numerical examples, we use Algorithm~\ref{alg:lambda}
to construct the values of $\lambda_{1},\dots,\lambda_{G}$.
For $\tBa \in \RR^{\abs{\cF}}$ and a subspace $S$ of $\RR^{\abs{\cF}}$, let $\dist(\tBa, S)=\min_{\Bs \in S}\norm{\tBa-\Bs}_{2}$.
In the step of $t=1$ of Algorithm~\ref{alg:lambda}, we choose the 1st design points $\Ba_{1}$ (setting $\lambda_{1}=0$ finally),
calculate the closeness from the subspace spaned by $\tBa_{1}$ for the other design points,
and choose one of the furthest design points from the subspace.
Next, in the step of $t=2$, we do the same procedure for the subspace spanned by the two chosen design points.
After the $\abs{\cF}$ iteration steps, each of $\lambda_{1},\dots,\lambda_{G}$ is determined as the accumulated sum of the closeness calculated in each step.
Note that
Algorithm~\ref{alg:lambda} is just one example
to determine the values of $\lambda_{1},\dots,\lambda_{G}$ destroying the symmetries,
and not necessarily desirable.
\begin{algorithm}[htbp]
 \makeatletter
 \renewcommand\p@enumii{}
 \makeatother
 \caption{An algorithm for constructing $\lambda_{1},\dots,\lambda_{G}$.}
 \label{alg:lambda}
\begin{enumerate}
 \item $t=1$, $N_{1}=\{1\}$.
 \item Repeat the following steps (\ref{alg:lambda:start})-(\ref{alg:lambda:end}) while $t \leq \abs{\cF}$.
       \begin{enumerate}
	\item \label{alg:lambda:start}
	      Let $S_{t}=\{\sum_{i \in N_{t}} \nu_{i}\tBa_{i} \in \RR^{\abs{\cF}} \mid \nu_{i} \in \RR, i \in N_{t}\}$
	      be a subspace of $\RR^{\abs{\cF}}$.
	\item Calculate $l_{gt}$ as follows:
	      \begin{eqnarray}
	       l_{gt} =
		\begin{cases}
		 0 & (g \in N_{t}) \\
		 \left\{\norm{\tBa_{g}}_{2}^{2}-(\dist(\tBa_{g}, S_{t}))^{2}\right\} & (g \notin N_{t})
		\end{cases}.
		\nonumber
	      \end{eqnarray}
	\item Select one of $g_{t+1} \in \{1,\dots,G\}\setminus N_{t}$
	      from the following set:
	      \begin{eqnarray}
	       \argmin_{g \in \{1,\dots,G\}\setminus N_{t}} \left\{\norm{\tBa_{g}}_{2}^{2}-(\dist(\tBa_{g}, S_{t}))^{2}\right\}.
		\nonumber
	      \end{eqnarray}
	\item \label{alg:lambda:end}
	      $N_{t+1}=\{g_{t+1}\} \cup N_{t}$, $t \leftarrow t+1$.
       \end{enumerate}
 \item Calculate 
	      $\lambda_{g} = \sum_{t=1}^{\abs{\cF}} l_{gt}$ 
       for $g=1,\dots,G$.
\end{enumerate}
\end{algorithm}

\begin{example}
 \label{example:selected:L4}
 As in Example~\ref{example:F3Level2:2}, we consider the case where $F=3$,
 and the model and model matrix are given by \eqref{eq:F3Level2:model-eq} and \eqref{eq:F3Level2:model-mat} respectively.
 Suppose that we want to estimate $\gamma_{1}, \gamma_{2}$ and $\gamma_{3}$.
 The values of the tuning parameters are
 $\lambda_{1}=\lambda_{4}=\lambda_{6}=\lambda_{7}=0$, $\lambda_{2}=\lambda_{3}=\lambda_{5}=\lambda_{8}=10$,
 where they are
 determined by Algorithm~\ref{alg:lambda}.
 By solving (\ref{eq:GroupLasso-SOCP-DOE}), we obtained the optimal $\Bbeta_{I_{g}}$'s.
 The computation time was 0.035 seconds.
 Table~\ref{table:selected:L4} is the optimal design matrix whose columns are the selected design points such that $\Bbeta_{I_{g}} \neq 0$.
 Note that Table~\ref{table:selected:L4} is equivalent to the $L_{4}$ orthogonal array.
\end{example}

\begin{table}[h]
 \caption{The optimal design matrix for the model of \eqref{eq:F3Level2:model-eq}}
 \label{table:selected:L4}
  \begin{center}
   \begin{tabular}{c|cccc}
    Run & $1$ & $2$ & $3$ & $4$ \\
    \hline
    $a_{1}$ & 1 & -1 & 1 & 1 \\
    $a_{2}$ & -1 & 1 & -1 & 1 \\
    $a_{3}$ & -1 & 1 & 1 & -1
   \end{tabular}
  \end{center}
\end{table}

\begin{example}
 \label{example:selected:L8}
 Assume that there are $F=4$ factors $a_{1}, a_{2}, a_{3}, a_{4}$ and each factor has two levels $-1$ or $1$.
 We consider all $G=2^{4}=16$ combinations of the levels of the factors as the candidate design points.
 Furthermore, we assume that the relation between the response variable and the factors is formulated as
 \begin{eqnarray}
  R
   =
  \gamma_{0}
  + \gamma_{1}a_{1} + \gamma_{2}a_{2} + \gamma_{3}a_{3} + \gamma_{4}a_{4}
  + \gamma_{5}a_{1}a_{2} + \gamma_{6}a_{1}a_{3} + \gamma_{7}a_{1}a_{4}
  + \epsilon.
   \nonumber \\
  \label{eq:F4Level2:L8:model-eq}
 \end{eqnarray}
 Suppose that we want to estimate $\gamma_{1}, \dots, \gamma_{7}$.
 Then, by solving (\ref{eq:GroupLasso-SOCP-DOE}) with the tuning parameters determined by Algorithm~\ref{alg:lambda},
 we obtained the optimal $\Bbeta_{I_{g}}$'s.
 The computation time was 0.163 seconds.
 Table~\ref{table:selected:L8} shows the optimal design matrix.
 Note that Table~\ref{table:selected:L8} is equivalent to the $L_{8}$ orthogonal array.
\end{example}

In the Example~\ref{example:selected:L4} and \ref{example:selected:L8},
the results are almost obvious
because we choose the values of the tuning parameters in order to obtain orthognal arrays
and only the runs whose values of $\lambda$ are zero are chosen in the optimal design matrix.
In the next example, we consider the case where a non-orthognal array is generated.

\begin{table}[h]
 \caption{The optimal design matrix for the model of \eqref{eq:F4Level2:L8:model-eq}}
 \label{table:selected:L8}
  \begin{center}
   \begin{tabular}{c|cccccccc}
    Run & $1$ & $2$ & $3$ & $4$ & $5$ & $6$ & $7$ & $8$ \\
    \hline
     $a_{1}$      & 1 &  1 &  1 &  1 & -1	& -1 & -1	& -1 \\
     $a_{2}$      & 1 &  1 & -1 & -1 &  1	&  1 & -1	& -1 \\
     $a_{3}$      & 1 & -1 & 	1 & -1 &  1	& -1 &  1	& -1 \\
     $a_{4}$      & 1 & -1 & -1 &  1 &  1	& -1 & -1	&  1 \\
     $a_{1}a_{2}$ & 1 &  1 & -1 & -1 & -1	& -1 &  1	&  1 \\
     $a_{1}a_{3}$ & 1 & -1 & 	1 & -1 & -1	&  1 & -1	&  1 \\
     $a_{1}a_{4}$ & 1 & -1 & -1 &  1 & -1	&  1 &  1	& -1
   \end{tabular}
  \end{center}
\end{table}
\begin{table}[h]
 \caption{The optimal design matrix for the model of \eqref{eq:F4Level2:L8plus:model-eq}}
 \label{table:selected:L8plus}
  \begin{center}
   \begin{tabular}{c|ccccccccc}
    Run & $1$ & $2$ & $3$ & $4$ & $5$ & $6$ & $7$ & $8$ & $9$ \\
    \hline
    $a_{1}$      & 1	&  1 &  1 &  1 & -1 & -1 & -1 & -1 & -1 \\
    $a_{2}$      & 1	&  1 & -1 & -1 &  1 &  1 &  1 & -1 & -1 \\
    $a_{3}$      & 1	& -1 & 	1 & -1 &  1 & -1 & -1 &  1 & -1 \\
    $a_{4}$      & 1	& -1 & -1 &  1 & -1 &  1 & -1 &  1 & -1 \\
    $a_{1}a_{2}$ & 1	&  1 & -1 & -1 & -1 & -1 & -1 &  1 &  1 \\
    $a_{1}a_{3}$ & 1	& -1 & 	1 & -1 & -1 &  1 &  1 & -1 &  1 \\
    $a_{1}a_{4}$ & 1	& -1 & -1 &  1 &  1 & -1 &  1 & -1 &  1 \\
    $a_{2}a_{3}$ & 1	& -1 & -1 &  1 &  1 & -1 & -1 & -1 &  1
   \end{tabular}
  \end{center}
\end{table}

\begin{example}
 \label{example:selected:L8plus}
 Assume that there are $F=4$ factors $a_{1}, a_{2}, a_{3}, a_{4}$ and each factor has two levels $1$ or $-1$.
 We consider all $G=2^{4}=16$ combinations of the levels of the factors as the candidate design points.
 Furthermore, we assume that the relation between the response variable and the factors is formulated as
 \begin{eqnarray}
  R
  & = &
  \gamma_{0}
  + \gamma_{1}a_{1} + \gamma_{2}a_{2} + \gamma_{3}a_{3} + \gamma_{4}a_{4}
  + \gamma_{5}a_{1}a_{2} + \gamma_{6}a_{1}a_{3} + \gamma_{7}a_{1}a_{4}
  \nonumber \\
  & &
  \qquad \qquad \qquad \qquad \qquad \qquad \qquad \qquad \qquad \qquad \qquad \quad
  + \gamma_{8}a_{2}a_{3}
  + \epsilon.
  \nonumber \\
  \label{eq:F4Level2:L8plus:model-eq}
 \end{eqnarray}
 Suppose that we want to estimate $\gamma_{1}, \dots, \gamma_{8}$.
 Then, by solving (\ref{eq:GroupLasso-SOCP-DOE}) with the tuning parameters determined by Algorithm~\ref{alg:lambda},
 we obtained the optimal $\Bbeta_{I_{g}}$'s.
 The computation time was 0.343 seconds.
 Table~\ref{table:selected:L8plus} shows the optimal design matrix.
 Note that the first $4$ columns of Table~\ref{table:selected:L8plus}
 are similar to Table~\ref{table:selected:L8}.
\end{example}

Finally, we give an example which shows how the computation time increases as the number of factors increases.
\begin{example}
  \label{example:computation_time}
  In this example, we observe the computation time changing the number of factors from $1$ to $9$.
  We assume that each factor has two levels $1$ or $-1$
  and the relation between the response variable and the factors is formulated as the main effect model (with no interaction terms) i.e.
  \begin{eqnarray*}
   R
    =
   \gamma_{0}
   + \gamma_{1}a_{1} + \gamma_{2}a_{2} + \cdots + \gamma_{F}a_{F}
   + \epsilon.
   \label{eq:computation_time}
 \end{eqnarray*}
  Table~\ref{tbl:computation_time} implies the computation time increases exponentially as the number of factors increases.
  \begin{table}
   \begin{center}
   \caption{The number of factors and computation time in seconds}
   \label{tbl:computation_time}
   \begin{tabular}{cc}
     $\#$ of factors & time [sec] \\ \hline
      1 & 0.021 \\
      2 & 0.029 \\
      3 & 0.037 \\
      4 & 0.136 \\
      5 & 0.325 \\
      6 & 1.301 \\
      7 & 6.987 \\
      8 & 81.172 \\
      9 & 742.403 \\
   \end{tabular}
   \end{center}
  \end{table}
\end{example}

\section{Concluding remarks}
\label{sec:concluding-remarks}

We apply the group lasso to the problem of constructing an optimal design matrix.
Though we mainly treat the case where each factor has two levels and
the set of candidate design points consists of all combinations of the levels of the factors without repetition
in the examples,
our approach, described in Section~\ref{subsec:DOE}, has the following features.
\begin{itemize}
 \item Each factor can have two or more levels.
 \item By duplicating the columns of the design matrix, we can treat the case of repeated measurements.
 \item By solving the problem of \eqref{eq:GroupLasso-Relaxation-DOE},
       we can obtain the optimal design matrix when we allow confounding among the factors.
 \item Assume that we have already observed the responses at the design points $\Ba_{g'_{1}}, \dots, \Ba_{g'_{Q}}$.
       Then, by setting $\lambda_{g'_{1}}=\dots=\lambda_{g'_{Q}}=0$,
       we can choose additional design points given $R_{g'_{1}}, \dots, R_{g'_{Q}}$.
\end{itemize}
Furthermore, we do not need to consider all combinations of the levels of the factors as the candidate design points.
If the number of the levels or the factors increases,
then the number of the combinations increases explosively.
This means that the number of the variables in the formulation of
\eqref{eq:GroupLasso-SOCP-DOE} or \eqref{eq:GroupLasso-Relaxation-DOE} increases explosively,
and hence it becomes difficult to solve the problem.
Therefore, if the number of the levels or the factors is large,
then it is needed to reduce the number of the candidates in advance to bound the number of variables
in \eqref{eq:GroupLasso-SOCP-DOE} or \eqref{eq:GroupLasso-Relaxation-DOE}.
For future work, we will investigate
how to choose the candidate design points in advance.
We will also investigate
how to determine the values of $\lambda_{g}$'s and $\kappa_{j}$'s.

\bibliographystyle{plainnat}
\bibliography{explasso}

\end{document}